\documentclass{article}

\usepackage{PRIMEarxiv}

\usepackage[utf8]{inputenc} 
\usepackage[T1]{fontenc}    
\usepackage{hyperref}       
\usepackage{url}            
\usepackage{booktabs}       
\usepackage{amsfonts}       
\usepackage{nicefrac}       
\usepackage{microtype}      
\usepackage{fancyhdr}       
\usepackage{graphicx}       
\graphicspath{{media/}}     
\usepackage{array}
\usepackage{makecell}
\setcellgapes{3pt}
\usepackage{enumitem}

\title{Bridging the Gap: Enhancing the Utility of Synthetic Data via Post-Processing Techniques
}

\author{
  Andrea Lampis, Eugenio Lomurno, Matteo Matteucci \\
  Politecnico di Milano \\
  Milan, Italy\\
  \texttt{\{andrea.lampis, eugenio.lomurno, matteo.matteucci\}@polimi.it} \\
}

\begin{document}
\maketitle


\begin{abstract}
Acquiring and annotating suitable datasets for training deep learning models is challenging. This often results in tedious and time-consuming efforts that can hinder research progress. However, generative models have emerged as a promising solution for generating synthetic datasets that can replace or augment real-world data. Despite this, the effectiveness of synthetic data is limited by their inability to fully capture the complexity and diversity of real-world data.
To address this issue, we explore the use of Generative Adversarial Networks to generate synthetic datasets for training classifiers that are subsequently evaluated on real-world images. To improve the quality and diversity of the synthetic dataset, we propose three novel post-processing techniques: Dynamic Sample Filtering, Dynamic Dataset Recycle, and Expansion Trick. In addition, we introduce a pipeline called Gap Filler (GaFi), which applies these techniques in an optimal and coordinated manner to maximise classification accuracy on real-world data.
Our experiments show that GaFi effectively reduces the gap with real-accuracy scores to an error of 2.03\%, 1.78\%, and 3.99\% on the Fashion-MNIST, CIFAR-10, and CIFAR-100 datasets, respectively. These results represent a new state of the art in Classification Accuracy Score and highlight the effectiveness of post-processing techniques in improving the quality of synthetic datasets.
\end{abstract}


\section{Introduction}\label{sec:introduction}
Over the last few years, deep generative models have become so powerful that they are able to produce high-quality samples that are almost indistinguishable from the real ones. With these recent developments, it is natural to ask whether these models are powerful enough to generate data that can be effectively used to train a machine learning model to perform a specific downstream task, thus completely replacing the real data function. This would have several advantages, for example it could significantly reduce the cost and effort of data collection, or it could be helpful in cases where information cannot be shared directly for privacy or sensitivity reasons, or where the original dataset is too large and the generative model can be used as a compressed version of the real data. 

In light of these considerations, and although the focus of these models has historically been on the perceptual quality of the data they generate, there have been attempts in recent years to formalise and quantify the usefulness of the synthetic data.
An essential contribution was made by Ravuri \textit{et al.}, who pioneered the metric called Classification Accuracy Score (CAS)~\cite{ravuri2019classification}. Given a system for generating data, the CAS represents the accuracy performance that a classifier trained solely on its generated data is able to achieve on a test set consisting instead of real data.
Surprisingly, despite the high perceptual quality of the data generated by the latest deep learning models, and despite the ability to generate an almost unlimited number of samples, training a model on them leads to a lower CAS value than the accuracy of the same model trained on real data.

In this paper, we investigate if and how we can bridge the utility gap between synthetic data generated by generative models and real-world data. We analyse the post-processing techniques available in the literature and propose new ones to improve synthetic data quality. We then present a new post-processing pipeline, Gap Filler (GaFi), which can be applied to any generative model. GaFi combines the most effective post-processing techniques to achieve a significantly better generator, without the need to modify the model's architecture or learning technique.

The contributions of our work are as follows:
\begin{itemize}
\item We propose two improved post-processing techniques, namely Dynamic Sample Filtering and Dynamic Dataset Recycle, and a novel method called Expansion Trick.
\item We propose the GaFi pipeline, which consists of a set of post-processing techniques suitable for any generative model to maximise the CAS achieved with its generated data.
\item We demonstrate the effectiveness of the GaFi pipeline by obtaining empirical CAS results that approach the upper bound of real accuracy performance. This achievement sets a new state of the art in generating synthetic data for classification tasks.
\end{itemize}

\section{Related Works}\label{sec:related}
In the past decade, deep learning has seen a surge in the development of generative models that are capable of producing synthetic data with increasing similarity to real-world training data. Some of the key architectures that have contributed to this progress include Variational Autoencoders (VAEs)~\cite{kingma2013auto}, Generative Adversarial Networks (GANs)~\cite{goodfellow2014generative}, and Denoising Diffusion Probabilistic Models (DDPMs)~\cite{ho2020denoising}.
These models have been predominantly used in the field of computer vision, particularly for image generation. To measure the perceptual quality of generated images, various metrics have been proposed, of which the Inception Score (IS)~\cite{salimans2016improved} and the Frécht Inception Distance (FID)~\cite{heusel2017gans} are the most significant.

In addition, the creation of synthetic datasets, either to replace or to complement real-world ones, has gained increasing attention in machine learning applications. Synthetic data can offer significant advantages, such as the possibility to generate large-scale datasets with known properties, reducing the need for costly data collection and annotation, and overcoming issues related to data privacy and access.
The use of generative models for synthetic data generation has found applications in several fields, including semantic segmentation~\cite{baranchuk2021label,li2022bigdatasetgan, li2021semantic, sankaranarayanan2018learning}, optical flow estimation~\cite{fischer2015flownet, NEURIPS2022_e8507db8, sun2021autoflow}, human motion understanding~\cite{varol2017learning, ma2022pretrained, izadi2011kinectfusion, 9879578}, and image classification~\cite{shmelkov2018good,ravuri2019classification,besnier2019dataset,lomurno2021sgde}.

A fundamental contribution to the advancement of this alternative use of generative models has been made by Ravuri \textit{et al.}, which focused on evaluating the performance of GANs through a downstream classifier, introducing the Classification Accuracy Score (CAS) metric~\cite{ravuri2019classification}. The underlying idea is to train a classifier with synthetic images and evaluate its performance on a test set composed of real images. The challenge behind this proposal is that if a generative model captures in an optimal way the real data distribution, performance on the downstream task should be similar whether using the original or synthetic data. 
Unfortunately, achieving comparable performance between classifiers trained on real and synthetic data remains a challenge. Significant efforts have been made to bridge this accuracy gap. 

One notable approach is the \textbf{Sample Filtering} technique proposed by Dat \textit{et al.}~\cite{8877479}, which aims to optimise the quality of the data generated. The technique uses an auxiliary classifier trained on real data to predict the labels of synthetic samples. Samples with incorrect predictions or those with low prediction confidence are discarded. In addition, the authors propose the use of \textbf{multiple generative models} to improve the accuracy of synthetic data by better capturing the real data distribution.
Another approach to address the accuracy gap is the \textbf{Dataset Smoothing} technique proposed by Besnier \textit{et al.}~\cite{besnier2019dataset}. This technique aims to create a diverse but gradually changing dataset by replacing only a portion of the generated training data with new samples at each epoch.

\section{Method}\label{sec:proposed_approach}
The present study aims to execute a comprehensive post-processing pipeline that can be applied to a broad range of generative models with the intention of improving their Classification Accuracy Score (CAS). To this end, we have meticulously surveyed the existing literature to identify the most effective techniques and subsequently adapted them to enhance their dynamicity and flexibility. In addition, we have devised and integrated novel methods into the proposed optimisation pipeline.

\subsection{Post-processing Techniques}
\label{subsec:post_processing_techniques}

\textbf{Dynamic Sample Filtering} Following Dat \textit{et al.} findings, we have re-implemented and extended their proposed technique using an adaptive approach that considers the dataset and generative model in use. As demonstrated in their ablation study, varying the filtering threshold may result in a synthetic dataset of superior quality compared to static filtering~\cite{8877479}.
To this end, we introduce Dynamic Sample Filtering, a two-step technique. Firstly, we utilize a classifier to predict the generated samples, thereby discarding all incorrectly classified samples. Secondly, we define a range of filtering thresholds, incrementally increasing from 0 to 0.9. For each threshold, we construct a standalone dataset consisting of only the correctly predicted samples with confidence greater or equal to the threshold.
We generate new data until the filtered samples reach the desired amount for each synthetic dataset. Finally, we train a classifier for each dataset and save the threshold value of the one achieving the best CAS. This technique helps eliminate low-quality images that could negatively impact the performance of the downstream classifier.

\par\medskip
\noindent\textbf{Dynamic Dataset Recycle} Inspired by Besnier \textit{et al.}, we propose to extend their Dataset Smoothing technique, which has shown significant benefits in their research. Our approach, called Dynamic Dataset Recycle, differs from the original in that it replaces the entire synthetic dataset in each iteration, rather than just a portion.
Our ablation studies show that recycling the entire dataset leads to better performance in terms of CAS.
Furthermore, to address the issue of time complexity, which is proportional to the size of the generated dataset, we propose to generalise its use by recycling the dataset every N epochs of classifier training.

\par\medskip
\noindent\textbf{Expansion Trick} We present a new method, the Expansion Trick, which works in contrast to the Truncation Trick proposed by Brock \textit{et al.}~\cite{brock2018large}. Instead of truncating the input noise space, we expand it by sampling from a normal distribution with a higher standard deviation than that used in model training. By increasing the diversity of the input noise space, our method encourages the generative model to explore regions that were less sampled during training. This leads to the production of more diverse and novel images, a desirable outcome in settings where diversity is prioritised over visual quality.
However, as expected, the increased standard deviation of the input noise distribution reduces the quality of individual samples. Therefore, the Expansion Trick is most effective when combined with sample filtering techniques. This mitigates the negative effects of lower sample quality by selecting only the most relevant samples to train the classifier.

\begin{figure}[t]
    \centering
    \includegraphics[width=1.\textwidth]{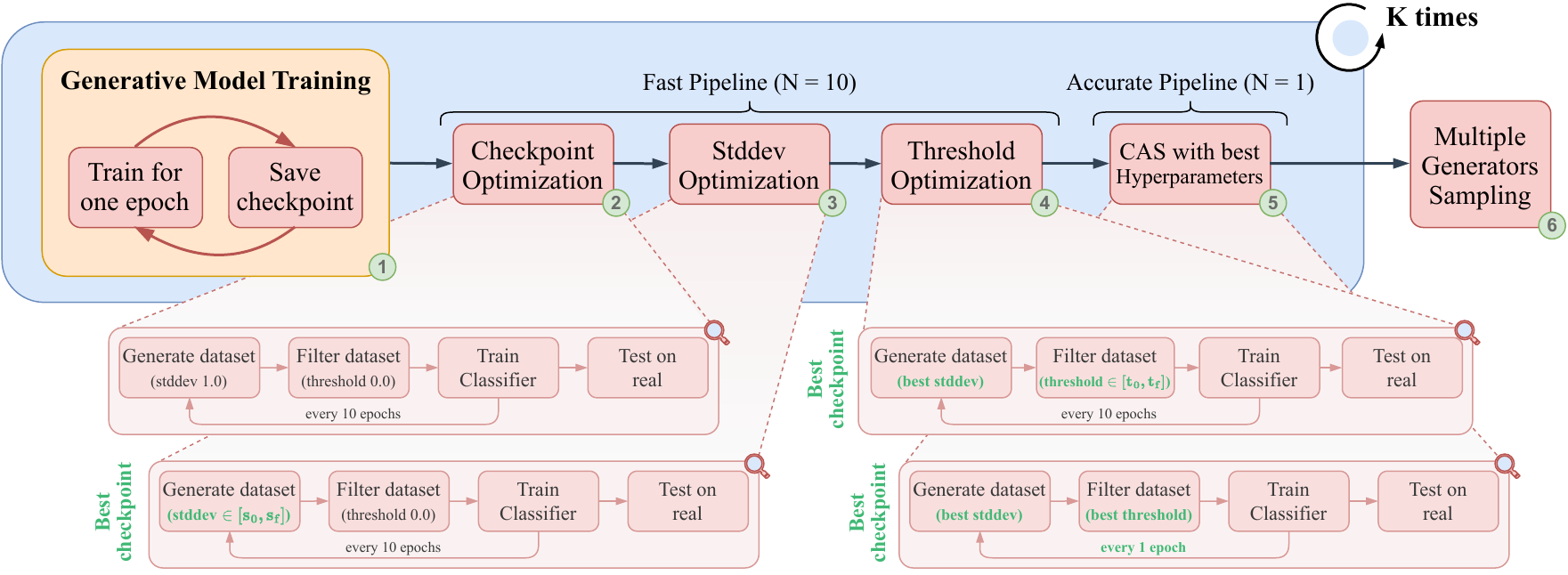}
    \caption[Proposed Pipeline]{Overview of the Gap Filler (GaFi) pipeline.}
    \label{fig:proposed_pipeline}
\end{figure}

\subsection{Gap Filler Pipeline}\label{subsec:pipeline}
In this section, we introduce the Gap Filler (GaFi) pipeline, which combines the post-processing techniques presented in the previous section. We discuss the importance of the correct application order and optimal hyperparameters for the techniques to work effectively and in synergy. The GaFi pipeline, depicted in Figure~\ref{fig:proposed_pipeline}, is composed of the following sequential steps:

\begin{enumerate}
    \item \textbf{Generative Model Training}: the initial step of the pipeline entails training a generative model and saving its checkpoints after every epoch for subsequent use in the pipeline. The specific type of generative model is not a constraint.
    \item \textbf{Checkpoint Optimization}: in order to optimise the performance of downstream classifiers, it is necessary to choose the best generative model among the saved checkpoints obtained during training. To accomplish this step, we propose evaluating each checkpoint by computing the CAS, and selecting the one that yields the highest performance. At this stage, we adopt a pipeline with fixed hyperparameters: a standard deviation $\mathit{stddev}$ of $1.0$ and a filtering threshold $\mathit{threshold}$ of $0.0$, meaning only samples predicted as the wrong class are discarded, without considering the confidence of the prediction. To balance the training time of the classifier, we set the Dataset Recycle technique parameter $N$ to $10$, which generates a new synthetic dataset every 10 epochs. This configuration is referred to as the \textit{"Fast Pipeline"}.
    \item \textbf{Stddev Optimization}: after identifying the best generative model checkpoint, we proceed to tune the hyperparameters. We start by determining the optimal standard deviation for the input noise distribution, which corresponds to the best configuration for the Expansion Trick. We achieve this by using the "Fast Pipeline" to calculate the CAS while varying the standard deviation between two predefined values, $s_0$ and $s_f$. In our experiments, we set $s_0=1.0$ and $s_f=2.0$, with incremental steps of $0.05$.
    \item \textbf{Threshold Optimization}: once the optimal standard deviation has been determined, the next step is to find the optimal filtering threshold for the Dynamic Sample Filtering technique. We follow a similar approach as before, using the "Fast Pipeline" to compute the CAS while varying the filtering threshold between two specified values, $t_0$ and $t_f$. In our experiments, we set $t_0=0.0$ and $t_f=0.9$, with increments of $0.1$.   
    \item \textbf{CAS with best hyperparameters}: finally, with the optimal hyperparameters in hand, we can proceed to train the classifier using the \textit{"Accurate Pipeline"}. Here, the Dynamic Dataset Recycle technique is set to $N=1$, allowing the use of data with a high degree of diversity to obtain the optimal classifier with respect to the generator under consideration.
    \item \textbf{Multiple Generators Sampling}: the final step of the GaFi pipeline is to create multiple generative models to sample the data and train a single optimal classifier. This is achieved by repeating all the previous steps of the pipeline $K$ times. According to Dat \textit{et al.}~\cite{8877479}, it is sufficient to train multiple identical generative models with different initialisation seeds on the same dataset. In this way, different aspects of the distribution of the dataset can be captured. A synthetic dataset is then created by sampling uniformly from these multiple models for each training epoch of the classifier.
\end{enumerate}

\section{Experiments Setup}\label{sec:setup}
In order to increase the transparency and reproducibility of the study, this section provides a comprehensive description of the experiments conducted and their setup. 
We selected the BigGAN Deep architecture as the generative model for our study~\cite{brock2018large}. 
To implement this architecture, we adopted the StudioGAN library~\cite{kang2022StudioGAN}, which makes slight modifications to the layer layout of the generator and discriminator residual blocks. Both $G$ and $D$ networks are initialised using the Orthogonal Initialization~\cite{saxe2013exact} technique, and trained using the Adam optimizer~\cite{kingma2014adam} with hyperparameters $\beta_1 = 0.5$, $\beta_2 = 0.999$, and a constant learning rate of $2e^{-4}$. 
We also utilized the Exponential Moving Average (EMA) technique for the weights of $G$ with a decay rate of 0.9999, as recommended by Brock \textit{et al.}~\cite{brock2018large}. Data augmentation was limited to random horizontal flipping of the training set. We trained all models using a batch size of 192 and with 3 $D$ steps per $G$ step.

We have chosen to employ the ResNet-20 architecture~\cite{he2016deep} as the downstream classifier due to its well-established performance. ResNet-20's width was set to 64, and the conventional ResNet training techniques were employed. This includes training it with cross-entropy loss, using a batch size of 128, training for 100 epochs, and using the SGD optimizer with an initial learning rate of 0.1, momentum of 0.9, and weight decay of $1e^{-4}$. The learning rate was reduced by a factor of 10 at epochs 60 and 80. To augment the synthetic training set, which has the same cardinality as the real dataset, a simple form of data augmentation was used. This involves zero-padding the input image, or its horizontally flipped version, to a size of $40\times40$, extracting a random crop of size $32\times32$, and using it as the final input image.

All experiments were conducted on three datasets, namely \texttt{Fashion-MNIST}~\cite{xiao2017fashion}, \texttt{CIFAR-10}~\cite{krizhevsky2009learning} and \texttt{CIFAR-100}~\cite{krizhevsky2009learning}. The \texttt{Fashion-MNIST} dataset contains 60,000 grey-scale $28\times28$ training images divided into 10 classes, while the other two datasets contain 50,000 RGB $32\times32$ training images divided into 10 and 100 categories, respectively. All three datasets have a test set of 10,000 images, which is used as the evaluation set for the classifiers. To make the images of the \texttt{Fashion-MNIST} dataset the same size as the other datasets, we resized them to $32\times32$ using zero padding.
The experiments have been conducted on a machine equipped with an Intel(R) Xeon(R) Gold 6238R CPU @ 2.20GHz CPU and an Nvidia Quadro RTX 6000 GPU. Training a single ResNet-20 model takes between 1 and 2.5 hours, depending on which and how many post-processing techniques are used, while training a BigGAN Deep requires around 48 hours.

\section{Results and Discussion}\label{sec:results}
\begin{table}[t]
    \caption[Filtering Threshold - Ablation]{The results of the CAS metric obtained using the Dynamic Sample Filtering technique for each filtering threshold.}
    \centering
    \resizebox{\textwidth}{!}{%
    \begin{tabular}{p{0.23\linewidth} >{\centering\arraybackslash}p{0.14\linewidth} *{10}{>{\centering\arraybackslash}p{0.0725\linewidth}}}
    \toprule
      & No Filtering & 0.0 & 0.1 & 0.2 & 0.3 & 0.4 & 0.5 & 0.6 & 0.7 & 0.8 & 0.9 \\
    \midrule
    \large\textbf{Fashion-MNIST} & \small88.70\% & \small89.88\% & \small90.01\% & \small89.59\% & \small89.98\% & \small89.89\% & \small90.05\% & \small\textbf{90.21}\% & \small89.86\% & \small90.12\% & \small89.90\%\\
    \large\textbf{CIFAR-10} & \small87.11\% & \small88.45\% & \small88.95\% & \small88.75\% & \small88.45\% & \small88.67\% & \small\textbf{89.06\%} & \small88.72\% & \small88.96\% & \small88.09\% & \small88.41\%\\
    \large\textbf{CIFAR-100} & \small57.74\% & \small59.13\% & \small58.82\% & \small\textbf{59.39\%} & \small59.35\% & \small59.20\% & \small59.06\% & \small58.79\% & \small58.76\% & \small57.28\% & \small55.52\%\\
    \bottomrule
    \end{tabular}
    }
\label{table:filtering_threshold_ablation_table}
\end{table}

\begin{table}[t]
    \caption[Dataset Recycle - Ablation]{The results of the CAS metric obtained using the Dynamic Dataset Recycle technique for each considered recycle frequency.}
    \centering 
    \begin{tabular}{p{0.20\linewidth} >{\centering\arraybackslash}p{0.12\linewidth} *{3}{>{\centering\arraybackslash}p{0.1\linewidth}} @{}}
    \toprule
      & \small{No Recycle} & \small{N=10}  & \small{N=5} & \small{N=1}\\
    \midrule
    \small\textbf{Fashion-MNIST} & \small88.70\% & \small89.29\% & \small89.88\% & \small\textbf{90.16\%}\\
    \small\textbf{CIFAR-10} & \small87.11\% & \small89.72\% & \small90.25\% & \small\textbf{90.42}\%\\
    \small\textbf{CIFAR-100} & \small57.74\% & \small59.68\% & \small60.57\% & \small\textbf{61.38}\%\\
    \bottomrule
    \end{tabular}
\label{table:dataset_recycle_ablation_table}
\end{table}

\begin{figure}[!t]
    \centering
    \includegraphics[width=1.\textwidth]{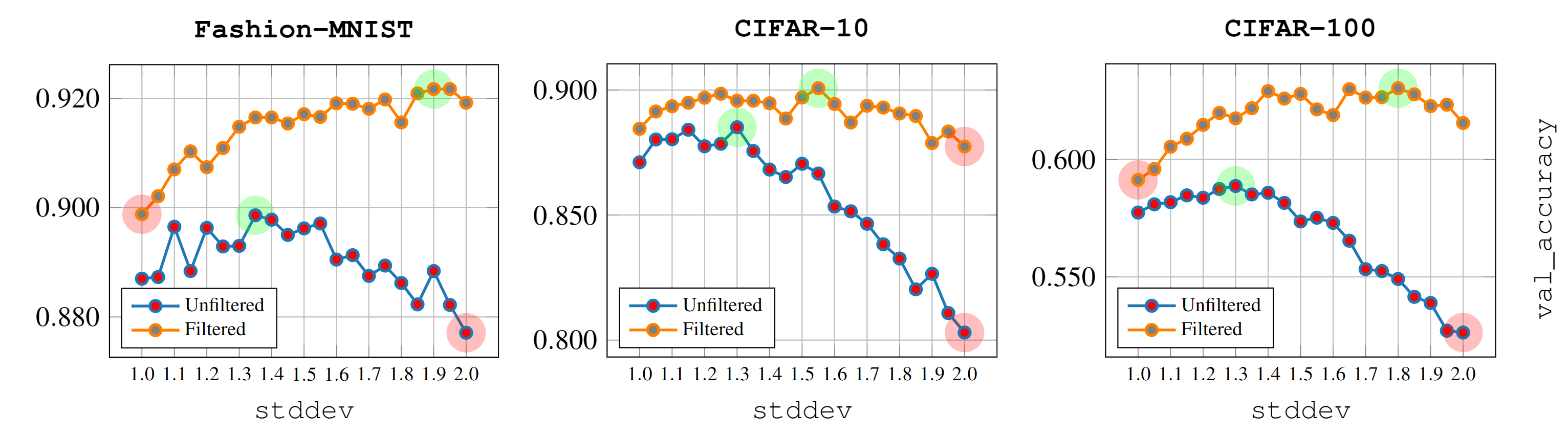}
    \caption[Expansion Trick - Ablation - Plots]{The results of the CAS metric obtained using the Expansion Trick technique. The plots compare unfiltered and filtered datasets (filtering threshold: 0.0).}
    \label{fig:expansion_trick_ablation_plots}
\end{figure}

In this section, we present the results of our experiments. We start by analysing the individual impact of each proposed post-processing technique. The results of these techniques are evaluated based on the final checkpoint of the generative model, i.e. without the application of the Checkpoint Optimization step. Finally, we present the results of the whole GaFi pipeline.

\par\medskip
\noindent\textbf{Dynamic Sample Filtering} As we can see from Table~\ref{table:filtering_threshold_ablation_table}, the use of the Sample Filtering technique is beneficial for all the datasets in analysis.
However, from this table it can be seen that the optimal threshold value is highly dependant on the specific dataset. For instance, the CAS for \textit{Fashion-MNIST} and \textit{CIFAR-10} remains almost constant for any threshold value, while for the \textit{CIFAR-100} it is clearly visible that a higher threshold value degrades the performance of the classifier. We assume that this behaviour is due to the fact that the generators trained on the first two datasets, being easier to learn, produce images that are very faithful to the original dataset. Therefore, the classifiers pretrained on real images will have high confidence in their predictions and most of the bad images will already be removed due to incorrect labelling. In contrast, \textit{CIFAR-100} is a much more complex dataset as it has 10 times the number of classes, causing the generated images to be more likely rejected by the pretrained classifier when a high filtering threshold is used. On average, with respect to the baseline CAS achieved, i.e. the "No Filtering" column of the table, the Dynamic Sample Filtering technique improves the CAS by 2\%.

\par\medskip
\noindent\textbf{Dynamic Dataset Recycle} Table~\ref{table:dataset_recycle_ablation_table} shows that the proposed dataset recycling technique significantly improves the CAS for all three datasets. The results reveal that even with a relatively soft recycling period, such as $N=10$, there is an increase in accuracy ranging from $0.59\%$ to $1.94\%$, depending on the dataset. Notably, by reducing the recycling period, i.e. generating new synthetic data more frequently during training, we can obtain an additional performance boost.
The gain in accuracy is more pronounced with increasing dataset complexity, as expected, since the generative model may require more attempts before generating meaningful data, especially for those classes learnt with poor effectiveness.

\par\medskip
\noindent\textbf{Expansion Trick} Figure~\ref{fig:expansion_trick_ablation_plots} displays the impact of the Expansion Trick on the CAS. The results indicate that when the dataset is unfiltered, there is a small increase in performance when using a standard deviation slightly higher than 1. However, as the standard deviation increases, there is a degradation in performance. This outcome is not unexpected, as a higher standard deviation leads to more diverse images but at the cost of image quality. As a result, beyond a certain point, the images become too degraded to be useful for training a downstream classifier. On the other hand, when using the Expansion Trick with a sample filtering technique, we can achieve significantly higher standard deviation values without compromising performance. This is because only the "good" images that meet the filtering criteria - in this case, the correctly classified ones - are kept, ensuring that the dataset is more diverse while still containing higher-quality images than the unfiltered dataset, leading to a higher classification accuracy.
Our novel technique improves the CAS by $3.47\%$, $2.96\%$, and $5.29\%$ on the \textit{Fashion-MNIST}, \textit{CIFAR-10}, and \textit{CIFAR-100} datasets, respectively. These gains demonstrate the efficacy of the Expansion Trick in enhancing the generative model's ability to produce informative samples, which in turn improves the downstream classifier's performance.

\par
\begin{figure}[t]
    \centering
    \includegraphics[width=1.\textwidth]{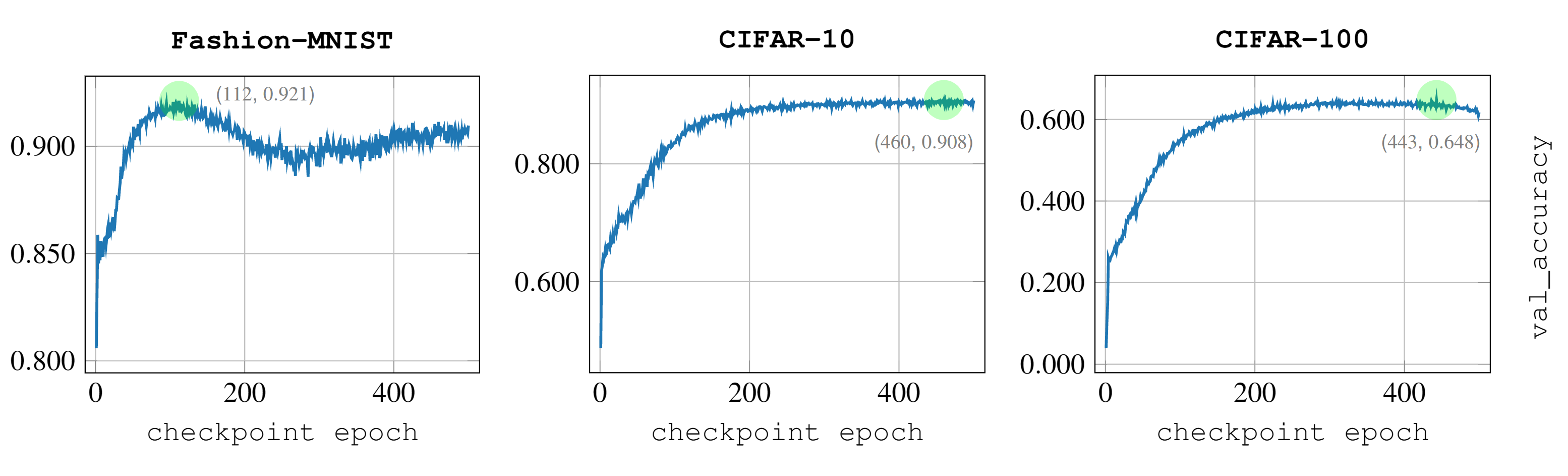}
    \caption[Expansion Trick - Ablation - Plots]{The results of the CAS metric obtained using the Expansion Trick technique. The plots compare unfiltered and filtered datasets (filtering threshold: 0.0).}
    \label{fig:checkpoint_optimization_plots}
\end{figure}

\medskip
\noindent\textbf{Checkpoint Optimization} The Figure~\ref{fig:checkpoint_optimization_plots} shows the evolution of the CAS metric as a function of the generative model checkpoint. Bearing in mind that each point corresponds to the complete training of a ResNet-20 classifier, the aim of this step is to identify the best checkpoint with respect to the CAS metric, even though it may be computationally expensive.
Looking at the graphs, the \textit{Fashion-MINST} dataset shows an optimal range of epochs between 90 and 120, while for the \textit{CIFAR-10} and \textit{CIFAR-100} datasets the optimal range is around epoch 450. It is noteworthy that the CAS increases with the number of epochs increases, up to a certain point. It is arguable that the complexity of the dataset is only partially indicative of the range in which to search for CAS, as can be assumed from the graphs of the \textit{CIFAR-10} and \textit{CIFAR-100} datasets.
In any case, this behaviour is consistent with that of GANs from the point of view of the perceptual quality of the generated images, which tend to collapse after a certain number of training iterations.
Overall, the Checkpoint Optimization step is fundamental and allows the next steps of the GaFi pipeline to proceed from the optimal generative model.

\begin{table}[t]
    \caption[CAS - Accurate Pipeline]{The optimal hyperparameters configuration and CAS performance obtained using the Accurate Pipeline.}
    \centering 
    \begin{tabular}{l c c c c}
    \toprule
     & \textbf{Checkpoint} & \textbf{Standard Deviation} & \textbf{Filtering Threshold} & \textbf{CAS}\\
    \midrule
    \textbf{Fashion-MNIST} & 112 & 2.00 & 0.0 & \textbf{94.03\%} \\
    \textbf{CIFAR-10} & 460 & 1.60 & 0.3 & \textbf{92.60\%} \\
    \textbf{CIFAR-100} & 443 & 1.70 & 0.1 & \textbf{68.92\%} \\
    \bottomrule
    \end{tabular}
    \label{table:cas_accurate_pipeline}
\end{table}

\par\medskip
\noindent\textbf{Accurate Pipeline} After determining the optimal configuration, resumed in Table~\ref{table:cas_accurate_pipeline}, the classifier can be retrained using the "Accurate Pipeline", where the recycling period $N$ is set to 1, regenerating the dataset at each training epoch to achieve the best possible CAS through the GaFi pipeline. It is evident that the Expansion Trick played a crucial role in achieving the optimal CAS in combination with the Dynamic Sample Filtering technique. This is supported by the heavily shifted values towards the standard deviation of 2 in each of the configurations.

The final results of the experiments are presented in Table~\ref{table:cas_our_vs_sota}, which includes the accuracy scores obtained from real data (\textit{Real Data}), the CAS of a single BigGAN Deep without post-processing techniques representing the baseline (\textit{Synthetic Baseline}), and for each number of generators considered, a comparison between our GaFi technique and the previous state-of-the-art post-processing performed by Dat \textit{et al.} using the same generative architecture used in this work.
Our approach achieves the best results, with improvements over the baseline of 5.33\% for \textit{Fashion-MNIST}, 6.09\% for \textit{CIFAR-10} and 14.21\% for \textit{CIFAR-100}.
Furthermore, our pipeline achieves higher accuracy even when using only one generator compared to the best configuration of Dat \textit{et al.} with six generators.
These results demonstrate that our proposed post-processing techniques, and the way they are applied in the GaFi pipeline, lead to obtaining superior classifiers trained on more generalised and useful data.

It is worth noting that the gap between our synthetic data and the real data has narrowed significantly. Specifically, for the \textit{Fashion-MNIST}, \textit{CIFAR-10} and \textit{CIFAR-100} datasets, the gap with respect to the baseline has been reduced from 7.31\%, 7.87\% and 17.9\% to 2.03\%, 1.78\% and 3.99\%, respectively. This remarkable result demonstrates the undeniable effectiveness of the GaFi pipeline. Moreover, it implies that the use of other generative models, whether existing or forthcoming, can further reduce this gap, and it may even be possible to achieve classifiers trained on synthetic data that outperform those trained on real data. This promising prospect illustrates the great potential of our proposed approach for synthesising high-quality data.

\begin{table}[t]
    \caption[CAS - Ours vs SotA]{The final results comparing the CAS obtained from the classifiers trained on generated data. The GaFi pipeline is compared with the previous state of the art, with the Synthetic Baseline and with the accuracy of the classifiers trained on real data.}
    \vspace{-5pt}
    \centering 
    \begin{tabular}{l l | l l l}
    \toprule
    \textbf{\#} & \textbf{Model} & \textbf{Fashion-MNIST} & \textbf{CIFAR-10} & \textbf{CIFAR-100}\\
    \toprule
    & Real Data & 96.01\% & 94.98\% & 75.64\%\\
    & Synthetic Baseline & 88.70\% & 87.11\% & 57.74\%\\
    \midrule
    \raisebox{-7pt}{1} & Dat \textit{et al.} Real Data & - & 88.25\% & 62.22\%\\
    & GaFi (ours) & 94.03\% & 92.60\%\ \scriptsize(+4.35\%)  &  68.92\%\ \scriptsize(+6.70\%)\\
    \midrule
    \raisebox{-7pt}{2} & Dat \textit{et al.} Real Data & - & 89.68\% & 64.33\%\\
    & GaFi (ours) & 93.98\% & 92.74\%\ \scriptsize(+3.06\%)  &  70.22\%\ \scriptsize(+5.89\%)\\
    \midrule
    \raisebox{-7pt}{4} & Dat \textit{et al.} Real Data & - & 90.68\% & 67.22\%\\
    & GaFi (ours) & 93.99\% & 93.02\%\ \scriptsize(+2.34\%)  &  71.75\%\ \scriptsize(+4.53\%)\\
    \midrule
    \raisebox{-7pt}{6} & Dat \textit{et al.} Real Data & - & 91.14\% & 67.56\%\\
    & GaFi (ours) & 93.98\% & 93.20\% \scriptsize(+2.06\%)  &  71.95\% \scriptsize(+4.39\%)\\
    \bottomrule

        \end{tabular}
    \label{table:cas_our_vs_sota}
\end{table}

\section{Conclusion}\label{sec:conclusion}
In this study, we introduced the Gap Filler pipeline (GaFi) to enhance the Classification Accuracy Score (CAS) by proposing new and enhanced post-processing techniques for generative models. These techniques included Dynamic Sample Filtering, Dynamic Dataset Recycle, and Expansion Trick, which have been shown to be highly beneficial when applied correctly. Our experimental results demonstrated that the proposed pipeline significantly increased the CAS, resulting in a new state-of-the-art performance on the three datasets analysed.
Despite our research yielding an accuracy that was slightly lower than that obtained on real data, we believe that the remaining gap raises a philosophical question about the very essence of generative modeling: whether it is possible to produce a model that can perfectly learn the distribution of real data. However, we remain optimistic that it is achievable.
We acknowledge that there are challenges that need to be addressed to bridge this gap, but we are confident that once this is achieved, it would open up new avenues for research and revolutionize several fields.

\section*{Acknowledgment}\label{sec:ack}
This project has been supported by AI-SPRINT: AI in Secure Privacy-pReserving computINg conTinuum (European Union H2020 grant agreement No. 101016577) and FAIR: Future Artificial Intelligence Research (NextGenerationEU, PNRR-PE-AI scheme, M4C2, investment 1.3, line on Artificial Intelligence).


\bibliographystyle{unsrt}  
\bibliography{references}

\end{document}